\documentclass[letterpaper, 10 pt, conference]{ieeeconf}  

\usepackage{amsmath,graphicx, hyperref}
\usepackage{booktabs}    
\usepackage{multirow}    
\usepackage{makecell}    
\usepackage{array}       
\usepackage{amsfonts}
\usepackage{amsmath,amssymb,amsfonts}
\usepackage{algorithmic}
\usepackage{graphics,graphicx,color}
\usepackage{textcomp}
\usepackage{xcolor}
\usepackage{algorithm}
\usepackage{algorithmic}
\usepackage{graphics} 
\usepackage{tabularx}
\usepackage{caption}

\IEEEoverridecommandlockouts                              

\title{\LARGE \bf
MulDP: Multimodal Diffusion Policy for Autonomous Quadruped Parkour Navigation across Complex Terrains
}

\author{\small Kangmai Hu$^{1}$, Yueqi Zhang$^{1}$, Peng Zhai$^{1,*}$, Xiaoyi Wei$^{1}$, Jiabin Hu$^{1}$, Zhixiang Liu$^{1}$, Quancheng Qian$^{1}$ and Lihua Zhang$^{1,*}$
\thanks{$^{1}$Kangmai Hu, Yueqi Zhang, Peng Zhai, Xiaoyi Wei, Jiabin Hu, Zhixiang Liu, Quancheng Qian and Lihua Zhang are with the College of Intelligent Robotics and Advanced Manufacturing, Fudan University, Shanghai 200433, China
        {\tt\small \{kmhu24, zhangyq23, weixy23, jbhu23, zxliu24, qcqian24\}@m.fudan.edu.cn; \{pzhai, lihuazhang\}@fudan.edu.cn}}%
\thanks{* Corresponding Author.}
}

\let\oldtwocolumn\twocolumn
\renewcommand\twocolumn[1][]{%
\oldtwocolumn[{#1}{
\begin{center}
\vspace{-0.6cm}

\includegraphics[width=\textwidth]{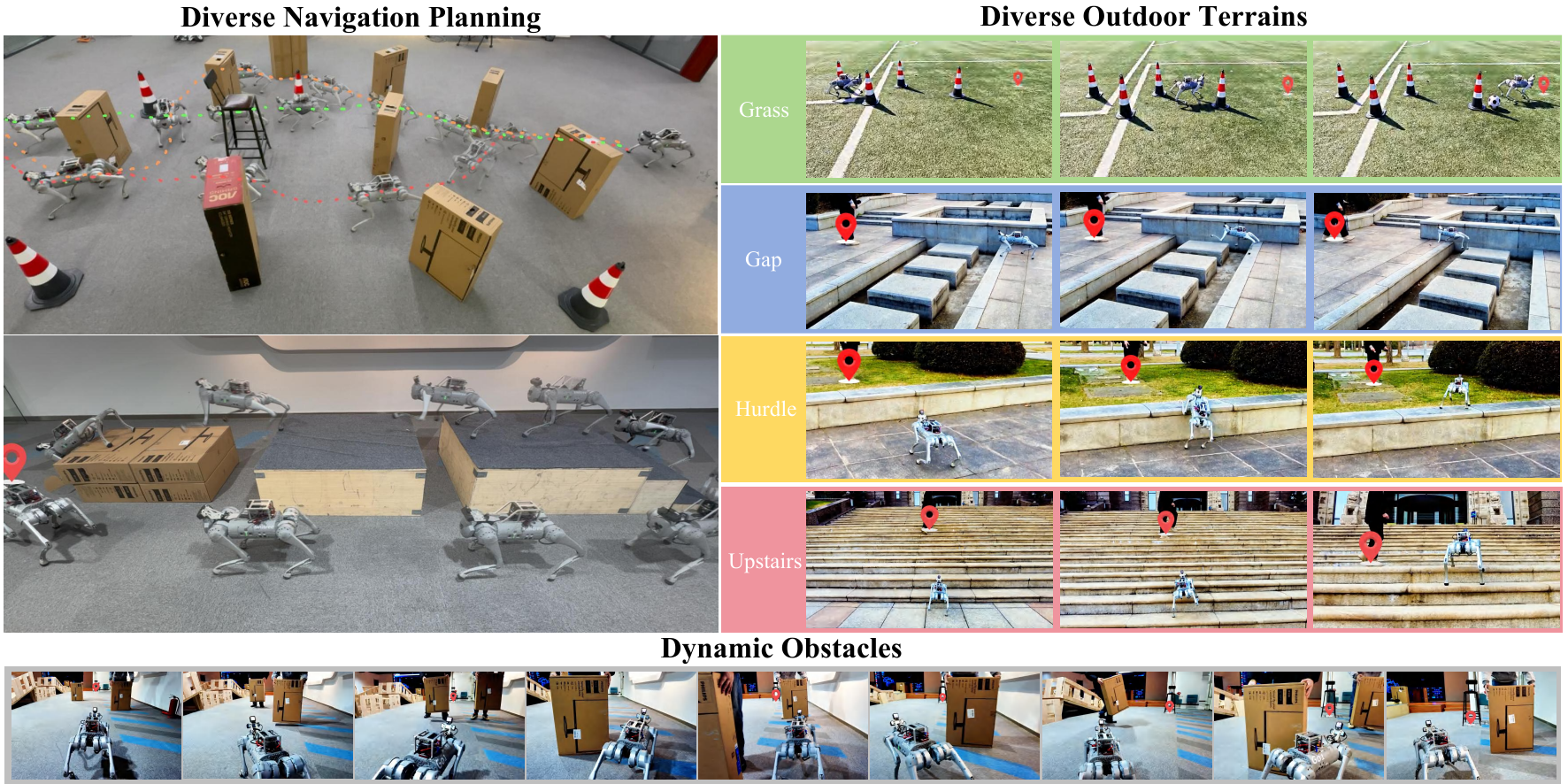}
\captionof{figure}{
\textbf{Left}: MulDP demonstrates diverse navigation behaviors in indoor environments with multiple feasible solutions, enabling the robot to autonomously leverage its locomotion capabilities to either traverse or circumvent obstacles.
\textbf{Right}: MulDP exhibits robust generalization across diverse outdoor environments, enabling efficient point-goal parkour navigation.
\textbf{Bottom}: MulDP performs real-time, safe autonomous navigation in dynamic obstacle scenarios.}
\label{headfig}
\vspace{-0.1cm}
\end{center}
    }]
}

\begin{document}
\maketitle
\thispagestyle{empty}
\pagestyle{empty}

\begin{abstract}

Quadruped robots have demonstrated impressive agility in parkour locomotion across complex terrains. However, most systems still rely on human intervention for high-level planning, and autonomous parkour navigation remains underexplored. The key challenges include fine-grained velocity regulation, long-horizon anticipatory behaviors, and tight coupling between perception and embodied execution. To address these challenges, we propose a Multimodal Diffusion Policy (MulDP) that integrates visual perception with robot proprioception and goal information to generate temporally coherent and anticipatory navigation velocity commands, tightly coupling perception with embodied control to enable robust autonomous navigation. To support the training of MulDP, we construct the first Quadruped Parkour Navigation Dataset (QPND), a multimodal dataset that encompasses diverse navigation behaviors and complex terrains. Extensive simulation and real-world experiments demonstrate that MulDP enables robust long-horizon autonomous navigation and effective traversal across complex terrains.

\end{abstract}

\section{INTRODUCTION}

Quadruped robots have achieved remarkable advances in mobility through learning-based methods \cite{extremeparkour, parkour3}, exhibiting strong agility and robustness across diverse terrains. Recent studies have enabled quadruped robots to execute parkour skills such as running, jumping, climbing, and vaulting, allowing them to traverse crowded streets \cite{NavOutdoor}, challenging terrains \cite{AMCO}, and cluttered warehouses \cite{GNM} to reach designated targets. Despite these advances in locomotion, existing work \cite{robotparkour} largely relies on human teleoperation for parkour navigation, leaving autonomous parkour navigation in complex environments insufficiently explored.

Compared to conventional navigation tasks that primarily focus on safe obstacle avoidance, autonomous robot parkour requires fine-grained temporal modulation of velocity in response to terrain variations to satisfy dynamic constraints and physical interaction demands. This is because parkour behaviors are inherently anticipatory and long-horizon: traversing a wide gap, for instance, requires accelerating in advance to accumulate sufficient momentum before reaching the edge. Moreover, visual traversability does not guarantee physical feasibility. The robot may avoid obstacles in egocentric view, yet still collide with them due to its body geometry and locomotion constraints.

Existing approaches struggle to address these challenges. Map-based navigation planners \cite{slam1, SLAM2, SLAM3} typically follow a modular design that separates mapping, planning, and control. Although effective in structured environments, such approaches may struggle in highly dynamic scenarios that require precise velocity modulation. Learning-based visual navigation methods \cite{iplanner, viplanner} bypass explicit mapping by predicting waypoints from visual observations, but often suffer from a decoupling between visual planning and physical execution, where obstacles are avoided within the camera’s field of view while the robot still experiences collisions in occluded or unobserved regions. End-to-end reinforcement learning \cite{AdSkill, skillnav, anymalparkour} and vision–language–action frameworks \cite{TrackVLA, VLA2} have been explored, but typically incur high training costs, exhibit limited generalization, and lack controllability in long-horizon tasks, further restricting their navigation performance in complex environments. 

In this work, we propose a \textbf{Multimodal Diffusion Policy (MulDP)} for autonomous quadruped parkour navigation. By formulating navigation velocity generation as a conditional generative diffusion process, MulDP addresses the intrinsic multimodality, long-horizon dependencies, and perception–control coupling challenges in autonomous parkour navigation. This framework enables real-time terrain-aware reasoning and dynamically consistent navigation behavior generation, improving robustness in complex terrains. Moreover, MulDP can serve as a local planner that is compatible with global planners, supporting more complex task-oriented navigation. Our contributions are as follows:

\begin{itemize}
\item We develop a novel multimodal diffusion policy that integrates historical visual observations, proprioception, and goal information, and iteratively denoises future navigation commands for temporally coherent and anticipatory long-horizon navigation in complex terrains.

\item We construct the first multimodal \textbf{Quadruped Parkour Navigation Dataset (QPND)} in simulation under sensing constraints. We collect quadruped parkour navigation behaviors in complex terrains using solely onboard perception and apply data augmentation to improve diversity, facilitating sim-to-real transfer.

\item We validate the effectiveness and robustness of MulDP through extensive evaluations in both simulation and real-world experiments.

\end{itemize}

\section{Related Work}
\subsection{Robot Diffusion Policy}

In practice, robot policy learning \cite{diffusionpolicy} poses unique challenges due to the inherently multimodal action distributions, strong temporal dependencies, and stringent precision requirements. Prior works using Gaussian mixture models \cite{GMM} or energy-based models \cite{IBC} remain limited in expressiveness and scalability, whereas diffusion models have been theoretically and empirically shown to approximate arbitrary distributions \cite{sde}, offering superior performance and stability in capturing complex multimodality.

Diffusion Policy \cite{diffusionpolicy} introduces observation-conditioned denoising diffusion probabilistic models for robotic manipulation, establishing a unified visuomotor framework that directly generates action sequences via iterative denoising conditioned on visual observations and robot states, rather than by direct regression. This formulation ensures stable training, smooth action generation, and robust performance across various manipulation and control tasks. Moreover, diffusion-based policies have been successfully extended to trajectory planning \cite{ldp}, data generation \cite{dategeneration}, dexterous manipulation \cite{dext}, and human–robot interaction \cite{human}, underscoring their broad applicability in robotics. Nevertheless, applying these models to autonomous quadruped parkour navigation remains challenging, due to both the scarcity of large-scale multimodal datasets and insufficient architectural adaptation to the task's temporal and embodied demands.

To address these issues, we incorporate three conditional encoders dedicated to historical perception modeling, fine-grained spatial representation, and goal conditioning into the diffusion policy, and construct a multimodal dataset in Isaac Sim with noise injection and data augmentation to facilitate sim-to-real generalization.

\subsection{Visual Navigation Models}
Visual navigation has been studied as a learning-based alternative to map-centric planning. Early approaches focus on waypoint or target navigation, where agents predict relative goal positions or intermediate waypoints from egocentric views. Point-goal navigation benchmarks demonstrate that visual inputs can effectively guide navigation without explicit maps. Recently, large-scale models \cite{GNM, ViNT} improve generalization by jointly learning visual representations and navigation policies across diverse environments. Several studies \cite{NavDP, NoMaD, dippest} have applied diffusion policies to visual navigation, enabling image-goal navigation, point-goal navigation, and goal-free navigation. These approaches exhibit promising obstacle avoidance performance and highlight the potential of generative diffusion models for visual navigation.

Despite their success, most visual navigation models are designed around waypoint-based planning, which limits their applicability to legged robots operating in complex three-dimensional terrains. Specifically, these methods \cite{viplanner} typically output waypoints, headings, or goal-relative vectors that are well suited for wheeled robots, but far from sufficient for agile legged locomotion requiring fine-grained velocity modulation and anticipatory control. Moreover, existing visual navigation approaches applied to quadruped robots largely overlook the robots’ inherent agility and treat navigation primarily as an obstacle avoidance problem, without explicitly leveraging locomotion capabilities for terrain traversal.

Compared with existing visual navigation models, our approach fuses visual perception with proprioceptive and other multimodal signals, and employs a generative diffusion policy to produce dynamically feasible and temporally coherent velocity commands. This design tightly couples perception and locomotion, enabling quadruped robots to fully exploit their locomotion capabilities in complex terrains.


\begin{figure*}[t!] 
    \includegraphics[
        width=\textwidth,
    ]{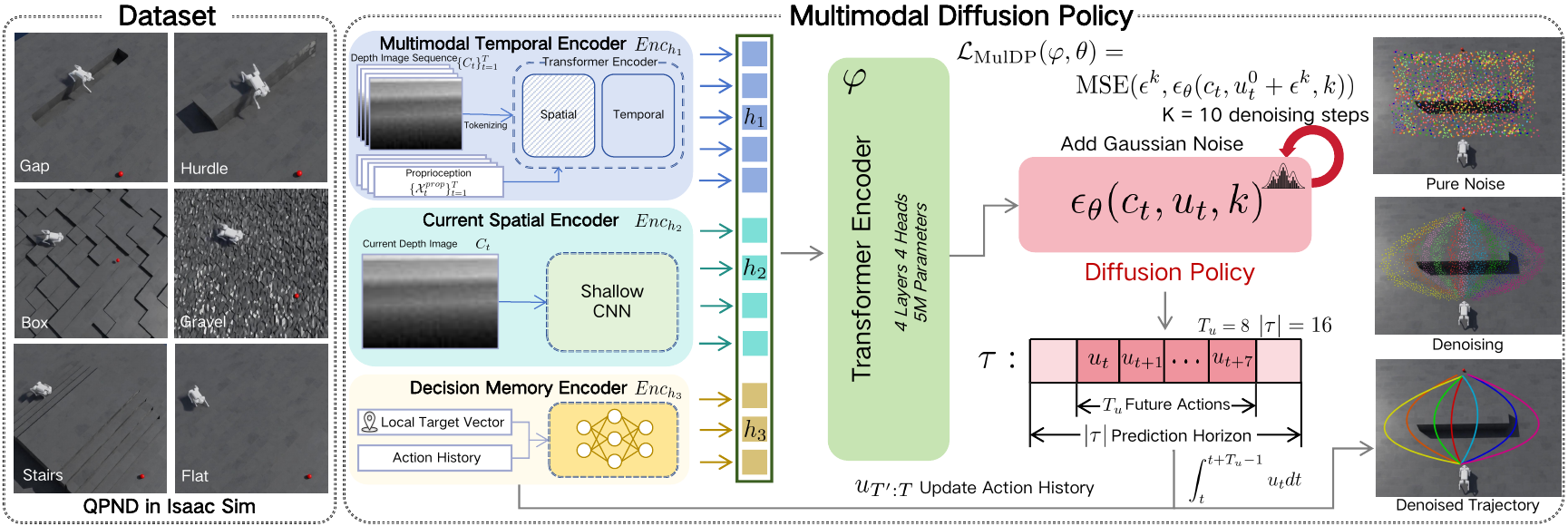}
    \caption{MulDP Architecture. Multimodal observations, including historical depth images, proprioception, the current depth frame, and goal information, are encoded and fused into a latent representation. The diffusion model conditions on this representation to iteratively generate a horizon of future velocity commands. At each 5 Hz update, only the first command in the predicted sequence is sent to the locomotion policy before the horizon is replanned.}
    \label{fig:system}
    \vspace{-0.4cm}
\end{figure*}

\section{Methodology}
\label{sec:Methodology}

\subsection{Overview}

For navigation tasks in complex environments, the MulDP system (as shown in Fig.~\ref{fig:system}) integrates depth images, proprioception, and decision memory to control locomotion. The navigation process is guided by a generative diffusion policy \cite{diffusionpolicy} that predicts precise control commands based on current and historical sensory inputs. With a diverse QPND dataset and effective modules, MulDP enables real-time, precise navigation in complex environments, ensuring smooth and adaptive locomotion. We detail the dataset generation in Sec.~\ref{data generation}, and provide a detailed discussion of the diffusion policy model and its training in Sec.~\ref{Diffusion policy} and Sec.~\ref{Model Training}.

\subsection{Data Generation}
\label{data generation}
In this section, we introduce the generation of the navigation dataset. Compared to previous works \cite{Gibson}, we define the dataset as a collection of depth images from the forward-facing camera, proprioception, and control commands. In contrast to previous quadruped approaches \cite{AdSkill, wheeledrobots} that assume access to local 3D information, our dataset requires only a forward-facing depth camera and proprioception, without introducing additional sensing devices such as LiDAR. This configuration preserves the non-privileged observations and the validity of the underlying interfaces, facilitating sim-to-real transfer. The dataset generation consists of five modules: (i) Robot Configuration, (ii) Reinforcement Learning Locomotion Policy, (iii) Scene Configuration, (iv) Trajectory Generation and (v) Data Augmentation.

\textbf{Robot Configuration.} We use Unitree Go1 in Isaac Sim and configure a RayCaster depth camera mounted on the robot's head with a fixed offset of [0.272, 0.0075, 0.092] meters relative to the body and rotated by a quaternion [0.9664, 0.0, 0.2571, 0.0]. The camera intrinsics are based on the calibration of a RealSense D435i. To account for pose uncertainties in real-world depth cameras, we randomly apply positional perturbations up to 10 mm and angular perturbations of $\pm 5^\circ$ during data collection. The camera updates at 10 Hz with a maximum range of 2 meters. This configuration ensures simulation efficiency while maintaining consistency with real depth sensor characteristics.

\textbf{Scene Configuration.} As illustrated in the dataset section of Fig.~\ref{fig:system}, we design terrains of varying difficulty levels in Isaac Sim, including gaps with widths ranging from 0.4 m to 0.7 m, hurdles with heights ranging from 0.4 m to 0.5 m, and stairs with individual step heights ranging from 0.15 m to 0.2 m. These terrains have feasible solutions within the solution set for traversing obstacles. In addition, we also include impassable obstacle terrains, such as obstacles with widths ranging from 0.3 m to 1.0 m and heights up to 1.0 m, to enhance the diversity of the scene.

\textbf{Reinforcement Learning Parkour Policy.} We employ a vision-based end-to-end reinforcement learning framework RENet \cite{RENet} that adjusts robot joint movements according to high-level control commands and visual inputs. Although RENet has been validated for its strong parkour performance, it lacks autonomous navigation planning capability. This policy also serves as the locomotion policy of our system.

\textbf{Trajectory Generation.} For basic terrains, we randomly sample 50 start–goal pairs and use an automated scripted policy (Scripted) to collect data. Specifically, fixed linear and angular velocity commands are applied to drive the robot toward the goal, while depth observations $C_t$, proprioception $\mathcal{X}_{t}^{prop}$, relative goal positions $g_t$, and control commands $u_t:(\Delta v_t, \Delta \theta_t)$ are recorded. The impassable obstacle terrains induce a multimodal action distribution with multiple feasible velocity profiles and traversal strategies. To adequately cover this diverse solution space, we rely on expert teleoperation (Teleop) to collect adaptive and multimodal traversal demonstrations. Table~\ref{dateset} provides a detailed configuration of the dataset. QPND covers a total navigation distance of 31.0 km, with 6,400 trajectories and approximately 250,000 depth images. It encompasses various complex terrains and multiple navigation solutions for the same terrain.

\textbf{Data Augmentation.}
To mitigate the sim-to-real gap and enhance the expressiveness of the learned policy, we apply additional processing during dataset collection.
\begin{itemize}
\item{Trajectory-level data augmentation.}
Considering the temporal continuity of navigation behaviors, we segment each collected trajectory into shorter contiguous motion sequences and treat each segment as an independent training sample. This improves data utilization by nearly 50\% while preserving the original expert behaviors.
\item{Sensor perception augmentation.}
To reduce the impact of edge noise in depth images on real-world deployment, we crop the boundary regions of depth images and rescale the central valid area. In addition, we inject 1\% random depth noise and Gaussian noise with a variance of 0.01 to simulate real-world measurement errors. Furthermore, noise is also added to the proprioception to alleviate the effects of sensor noise during deployment.
\end{itemize}

\begin{table}[htbp]
  \centering
  \footnotesize
  \captionsetup{font=small}
  \setlength{\tabcolsep}{3pt}
  \renewcommand{\arraystretch}{0.85}
  \vspace{-0.2cm}
  \caption{Composition of QPND.}
  \vspace{-0.1cm}
  \begin{tabular}{c c c c}
    \toprule
    \textbf{Terrain} & \textbf{Distance (km)} & \textbf{Trajectory (K)} & \textbf{Collection} \\
    \midrule
    \multirow{2}{*}{Gap}    & 2.1 & 0.7 & Scripted \\[-0.4ex]
    \cmidrule(lr){2-4}
                            & 6.8 & 1.1 & Teleop   \\
    \midrule
    \multirow{2}{*}{Hurdle} & 1.8 & 0.6 & Scripted \\[-0.4ex]
    \cmidrule(lr){2-4}
                            & 6.9 & 1.1 & Teleop   \\
    \midrule
    Stairs & 3.4 & 0.6 & Scripted \\
    \midrule
    Gravel & 1.5 & 0.5 & Scripted \\
    \midrule
    Box & 2.1 & 0.6 & Scripted \\
    \midrule
    Flat & 6.4 & 1.2 & Scripted \\
    \bottomrule
  \end{tabular}
  \label{dateset}
  \vspace{-0.2cm}
\end{table}

\subsection{Diffusion Policy}
\label{Diffusion policy}
The navigation module is implemented through a novel multimodal diffusion policy. Specifically, we introduce a Multimodal Temporal Encoder to model historical depth observations and proprioceptive signals, enabling explicit temporal context reasoning. We further design a Current Spatial Encoder to capture fine-grained spatial structure from the current observation. The encoded features are then fused with goal features into a unified latent representation, which conditions the iterative denoising process to generate a future navigation-command horizon. An overview of the MulDP architecture is shown in Fig.~\ref{fig:system}.

\textbf{Multimodal Temporal Encoder.} MulDP processes historical depth images \(\{C_t\}_{t=1}^T\) and proprioception \(\{\mathcal{X}_{t}^{prop}\}_{t=1}^T \) as inputs, constructing a unified spatiotemporal feature representation for robust environmental understanding. Specifically, we design a multimodal Transformer encoder that partitions 44$\times$56 depth image sequences \(C_t\) into 4$\times$4 patches (with 154 visual tokens), which are jointly embedded with 45-dimensional proprioception \(\ \mathcal{X}_{t}^{prop}\ \) into a unified feature space. Learnable modality embeddings distinguish different input sources, with all tokens undergoing cross-modal attention in the Transformer encoder to produce 128-dimensional fused conditioning features \(h_1=Enc_{h_1}(\{C_t, \mathcal{X}_{t}^{prop}\}_{t=1}^T)\).

\textbf{Current Spatial Encoder.} While the Multimodal Temporal Encoder captures historical visual and proprioceptive features during locomotion, complex terrain navigation requires refined topographic features. For example, when encountering sudden obstacles, relying solely on abstract features from \(h_1\) proves insufficient for safe navigation. Therefore, the second component introduces a parallel processing path dedicated to immediate perception. This component employs a shallow CNN to process the current depth image \(C_t\), generating detailed spatial features \(h_2=Enc_{h_2}(C_t)\).

\textbf{Decision Memory Encoder.} To improve locomotion planning continuity and intentionality, we incorporate a spatiotemporal decision memory module that stores the last 5 steps of planning controls and goal vectors. To generate smooth, goal-oriented navigation commands, an MLP models the geometric relationship between decision history and target variation into a latent feature \(h_3=Enc_{h_3}(\{u_{t-i}\}_{i=0}^4)\).

\textbf{Generative Diffusion Policy}. Our diffusion policy predicts the command horizon \(U_t=\{u_{t+i}\}_{i=0}^{H-1}\in\mathbb{R}^{H\times2}\). Each command is \(u_{t+i}=(\Delta v_{t+i},\Delta \theta_{t+i})\). First, we construct the conditional context \(c_t\) for the noise prediction network. The framework utilizes a 4-layer, 4-head architecture with 1 conditional layer to process input features from components \(h_1\), \(h_2\) and \(h_3\), fusing global conditions \(c_t\) for the DDPM scheduler \cite{DDPM} during the denoising process.

We initialize the future action horizon \(U_{t}^K\) from a Gaussian distribution and perform K denoising iterations to generate an intermediate sequence \(\{U_t^K, U_t^{K-1}, \dots, U_t^0\}\) with progressively reduced noise levels, ultimately producing the desired noise-free output horizon \(U_t^0\). The denoising process follows the equation:
\begin{equation}
U_t^{k-1} = \lambda \cdot \left( U_t^k - \mu \epsilon_\theta(c_t, U_t^k, k) \right) + \mathcal{N}(0, \sigma^2 I)
\end{equation}
where \(k\) represents the number of denoising steps, \(\epsilon_\theta\) is the noise prediction network parameterized by \(\theta\), and \(\lambda, \mu\) and \(\sigma\) are functions of the noise schedule. 

The noise prediction network \(\epsilon_\theta\) is conditioned on the observation context \(c_t\), integrating multimodal perception data. We set the number of denoising steps to 10 to balance output accuracy and inference speed. During deployment, only the first command in \(U_t^0\) is sent to the locomotion policy, and the full horizon is regenerated at the next 5 Hz update.

\subsection{Model Training}
\label{Model Training}
We train MulDP on QPND using the following loss function for end-to-end supervised learning: 
\begin{equation}
    \mathcal{L}_{\text{MulDP}}(\varphi,\theta) =\mathbb{E}_{\tau,C,\mathcal{X}_{prop},g \sim \mathcal{D}}[(\epsilon^k - \epsilon_\theta(c_{t}, U_{t}^0 + \epsilon^k, k))^2]
\end{equation}
where the expert trajectories $\tau$ without noise, depth images $C$, relative goals $g$, and proprioception \(\mathcal{X}_{prop}\) are sampled from the QPND dataset \(\mathcal{D}\), with a history size T of 5 for each modality.  \(\varphi\) corresponds to the Transformer layers and \(\theta\) corresponds to the diffusion process parameters. The diffusion strategy employs $K=10$ denoising steps and DDPM for training. We use the AdamW optimizer with a learning rate of \(10^{-4}\), train MulDP for 1000 epochs with a batch size of 512, and apply cosine scheduling and warm-up to stabilize the training process.

\section{Experiments}
\label{sec:exp}
In this section, we aim to answer the following questions using both quantitative and qualitative experimental results:
\begin{itemize}
    \item Q1: Do the individual components of MulDP each contribute to improved navigation and parkour capabilities? Moreover, what advantages does MulDP offer over existing quadruped robot navigation algorithms?
    \item Q2: Can MulDP demonstrate flexible and diverse fixed-goal navigation in challenging environments?
    \item Q3: Can MulDP perform robust and generalizable dynamic target-following navigation in complex outdoor environments?
\end{itemize}

\subsection{Experimental Setup and Deployment Details}
\label{exp-setup}
In simulation, we evaluate our method on fixed-goal navigation tasks using Unitree Go1. We construct a navigation evaluation benchmark based on Isaac Sim, covering six terrain types: Gap, Hurdle, Upstairs, Downstairs, Gravel, and Box. Each terrain type includes two levels of difficulty. Four metrics are considered: \textbf{Success Rate (SR)}, defined as reaching the target within 0.2 m; \textbf{Success weighted by Path Length (SPL)} \cite{evaluation}, which jointly measures navigation success and path efficiency; \textbf{Traversal Rate (TR)}, defined as the ratio of the area traversed by the robot within the valid navigation region; and \textbf{Time to Reach (TTR)}, which quantifies the average time required for the robot to reach the goal in successful trials. For each scenario, the robot’s initial position is randomized over 50 distinct locations, and goal positions are sampled at distances ranging from 5 m to 8 m. The maximum time limit for each navigation episode is set to 60 s.

For real-world experiments, we evaluate the performance of fixed- and dynamic-goal parkour navigation using MulDP in both indoor and outdoor environments. MulDP is deployed on a Unitree Go1 equipped with an NVIDIA Orin NX 16GB, operating at an inference frequency of 5 Hz. A RealSense D435i camera is used, and all sensor configurations are kept consistent with those used during training. To provide the relative position between the robot and the target, we employ a high-precision Ultra-Wideband (UWB) localization system.

\subsection{Comparison and Ablation Study}

\subsubsection{\textbf{Comparison Settings}} We compare MulDP with recent diffusion-based methods and point-goal navigation approaches. The test terrains are shown in Section~\ref{exp-setup}.
\begin{itemize}
    \item NavDP\textsuperscript{*}\cite{NavDP}: A critic-free version of NavDP trained on the QPND dataset. The original NavDP critic is trained and evaluated based on ESDF, which is specifically designed for collision-free navigation by penalizing proximity to obstacles. However, in parkour scenarios where obstacle traversal is allowed, such a criterion imposes overly conservative constraints.
    \item ViNT\textsuperscript{*}\cite{ViNT}, PointNav\textsuperscript{*}\cite{VLFM}: Models are trained on the QPND dataset for 100 epochs.
    \item NavDP, ViNT: We directly evaluate NavDP and ViNT using their officially released checkpoints, while keeping robot-related configurations consistent.
\end{itemize}

\begin{figure}[htbp]
\vspace{0.2cm}
\centerline{\includegraphics[width=\columnwidth]{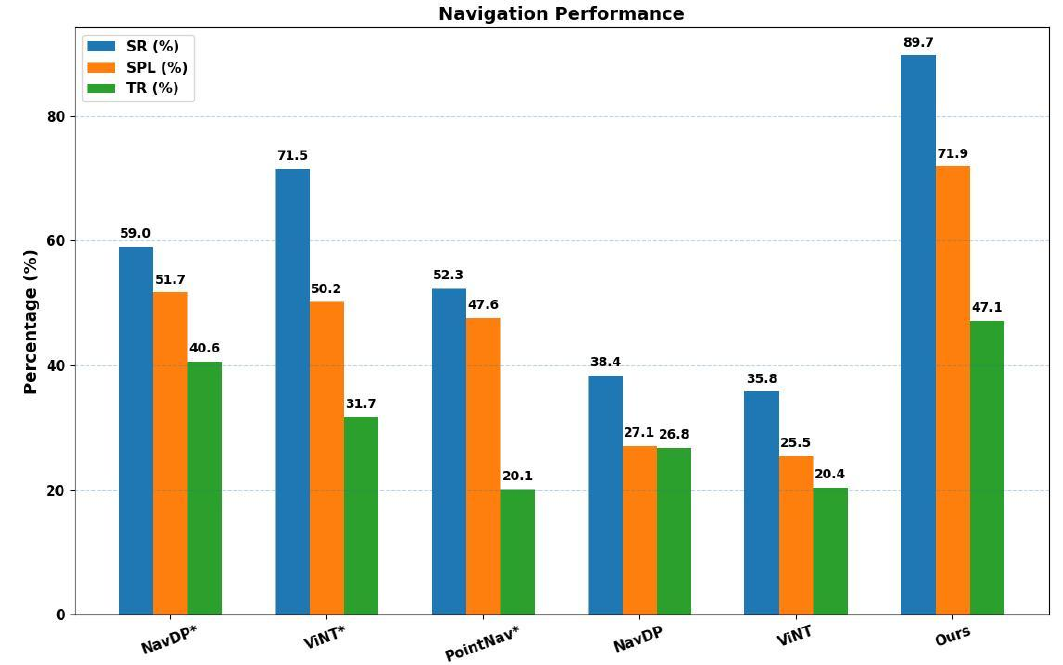}}
\caption{Comparison Experiments of MulDP. }
\label{compare}
\vspace{-0.2cm}
\end{figure}

We compare MulDP with a set of baseline methods. Compared to the diffusion-based approach NavDP\textsuperscript{*}, MulDP achieves a higher success rate (SR) of 89.7\%, compared to 59.0\% for NavDP\textsuperscript{*}. This performance gap arises because MulDP directly consumes forward-facing depth images as input and jointly trains the visual encoder and navigation policy end-to-end. In contrast, NavDP\textsuperscript{*} relies on a pre-trained DepthAnything encoder to extract features from RGB images, which introduces a representational mismatch between the pre-training domain and the navigation task. Compared with non-diffusion-based methods ViNT\textsuperscript{*} and PointNav\textsuperscript{*}, both MulDP and NavDP\textsuperscript{*} achieve traversal rates (TR) exceeding 40\%, highlighting the advantage of diffusion-based policies in navigation. Among the non-diffusion baselines, ViNT\textsuperscript{*} attains the highest SR of 71.5\%, which can be attributed to its joint training of the visual encoder and navigation policy, thereby mitigating errors introduced by decoupled perception and decision-making.

We further evaluate the checkpoints released by NavDP and ViNT. These checkpoints fail in high-difficulty terrain scenarios due to their inability to traverse obstacles, although they remain effective in obstacle-avoidance settings, with SR dropping to 38.4\% and 35.8\%, respectively. Their navigation solutions converge to conservative obstacle-avoidance behaviors and fail to exploit the robot’s locomotion capabilities to identify more efficient traversable paths. Consequently, their SPL scores decrease to 27.1\% and 25.5\%, respectively, indicating poor path efficiency.

Overall, MulDP achieves an average SR of 89.7\%, an average SPL of 71.9\%, and an average TTR of 4.9 s across all evaluated scenarios, demonstrating its superior ability to identify effective planning solutions and guide the robot through complex environments. 

\subsubsection{\textbf{Ablation Settings}} We perform experiments to evaluate the contribution of each module in our method.
\begin{itemize}
    \item w/o Prop: Proprioceptive inputs are removed, and only depth images are used as input.
    \item w/o DME: The Decision Memory Encoder is removed.
    \item w/o CSE: The Current Spatial Encoder is removed.
    \item w/o DN: The model is trained on a dataset without noise injection on depth images and proprioceptive signals.
    \item w/o DA: The model is trained on a dataset without trajectory augmentation.
\end{itemize}

\begin{table}[htbp]
  \centering
  \small
  \setlength{\tabcolsep}{3pt}       
  \renewcommand{\arraystretch}{1} 
  \captionsetup{font=small}
  \vspace{-0.2cm}
  \caption{Ablation Experiments of MulDP.}
  \vspace{-0.2cm}
  \begin{tabular}{c c c c c}
    \toprule
    Method & SR(\%) $\uparrow$ & SPL(\%) $\uparrow$ & TR(\%) $\uparrow$ & TTR(s) $\downarrow$ \\
    \midrule 
    w/o Prop & 69.5 & 45.8 & 34.3 & 7.5 \\
    w/o DME  & 7.4 & 0.1 & \textbf{90.3} & 58.4 \\
    w/o CSE  & 82.0 & 65.1 & 46.8 & 5.0 \\
    w/o DN & 75.0 & 62.6 & 35.5 & 8.0 \\
    w/o DA & 32.5 & 23.7 & 21.0 & 5.5 \\
    \midrule
    Ours & \textbf{89.7} & \textbf{71.9} 
         & 47.1 & \textbf{4.9} \\
    \bottomrule
  \end{tabular}
  \label{CAstudy}
\vspace{-0.2cm}
\end{table}

We conduct an ablation study to analyze the contribution of each component in MulDP, revealing that proprioception is indispensable for parkour navigation, while goal orientation reasoning is the ultimate navigation objective. As shown in Table~\ref{CAstudy}, removing proprioceptive inputs (w/o Prop) leads to a substantial performance degradation, with the success rate decreasing from 89.7\% to 69.5\% and the time to reach increasing from 4.9 s to 7.5 s. These results highlight the necessity of tightly coupling body-state awareness with visual perception in parkour navigation. When the decision memory encoder is removed (w/o DME), the SR decreases sharply to 7.4\%, while the TR increases to 90.3\%, indicating that the robot continues to explore the environment extensively without effectively converging toward the target. The current spatial encoder encodes the robot’s instantaneous visual perception. Removing this module (w/o CSE) leads to only a modest reduction in SR, from 89.7\% to 82.0\%, but can still induce erroneous navigation decisions in challenging dynamic scenarios, resulting in suboptimal planning.

In addition, both noise injection and data augmentation applied to the QPND dataset are essential for robust sim-to-real generalization. Data augmentation (DA) constitutes a core component of QPND by substantially increasing trajectory diversity and enhancing dataset generalization. Without DA (w/o DA), the dataset size is reduced by nearly 50\%, which exacerbates overfitting and severely degrades generalization performance, with SR dropping to 32.5\%, SPL to 23.7\%, and TR to 21.0\%. Training without data noising (w/o DN) causes MulDP to overfit to specific visual appearances in certain environments, leading to a reduced SR of 75.0\%. 

\begin{figure}[!b]
\centerline{\includegraphics[width=\columnwidth]{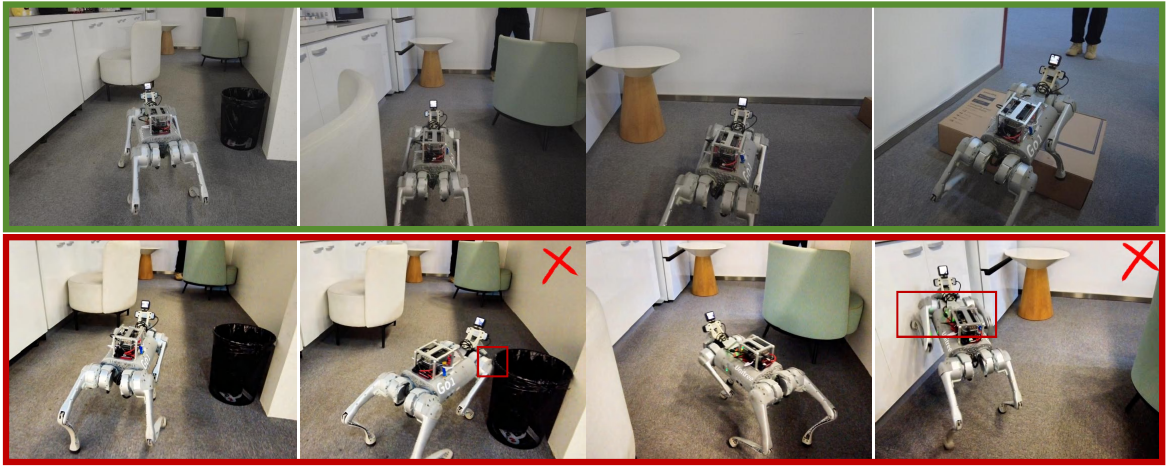}}
\caption{Performance comparison of MulDP and w/o DN in a narrow residential environment. \textbf{\textcolor[rgb]{0.345,0.557,0.192}{Green box}}: MulDP successfully navigates through the narrow corridor. \textbf{\textcolor[rgb]{0.753,0,0}{Red box}}: w/o DN model collides with obstacles during navigation.}
\label{cl}
\end{figure}

As shown in Fig.~\ref{cl}, we further validate the importance of dataset noise for sim-to-real transfer in a narrow residential indoor environment, where obstacle types and scene layouts are unseen during training. The deployed MulDP robot successfully perceives non-traversable obstacles and avoids collisions. When encountering a suddenly appearing box, the robot executes a climbing maneuver to traverse it. In contrast, due to inevitable depth sensing noise in real-world settings, the deployed w/o DN robot misjudges obstacle geometry, collides with it, and fails to distinguish between low and tall obstacles, resulting in inappropriate traversal behaviors. These results demonstrate that dataset noise is essential for effective sim-to-real generalization.

\subsection{Fixed-goal Navigation Task}

\begin{figure}[t!]
\vspace{0.2cm}
\centerline{\includegraphics[width=\columnwidth]{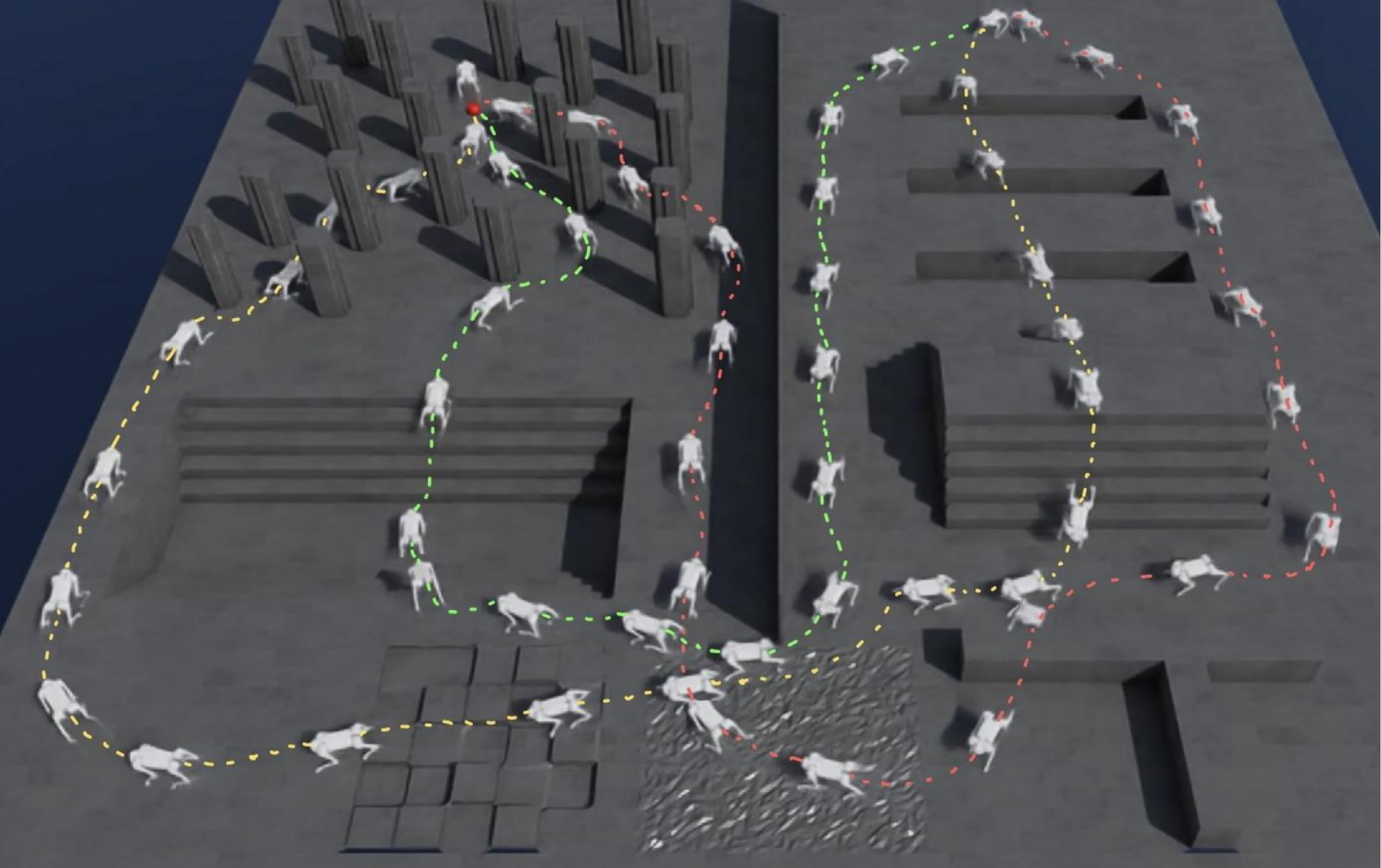}}
\caption{Long-horizon parkour navigation in Isaac Sim.}
\label{sim}
\vspace{-0.6cm}
\end{figure}

\begin{figure*}[t!] 
\vspace{0.2cm}
    \includegraphics[
        width=\textwidth,
    ]{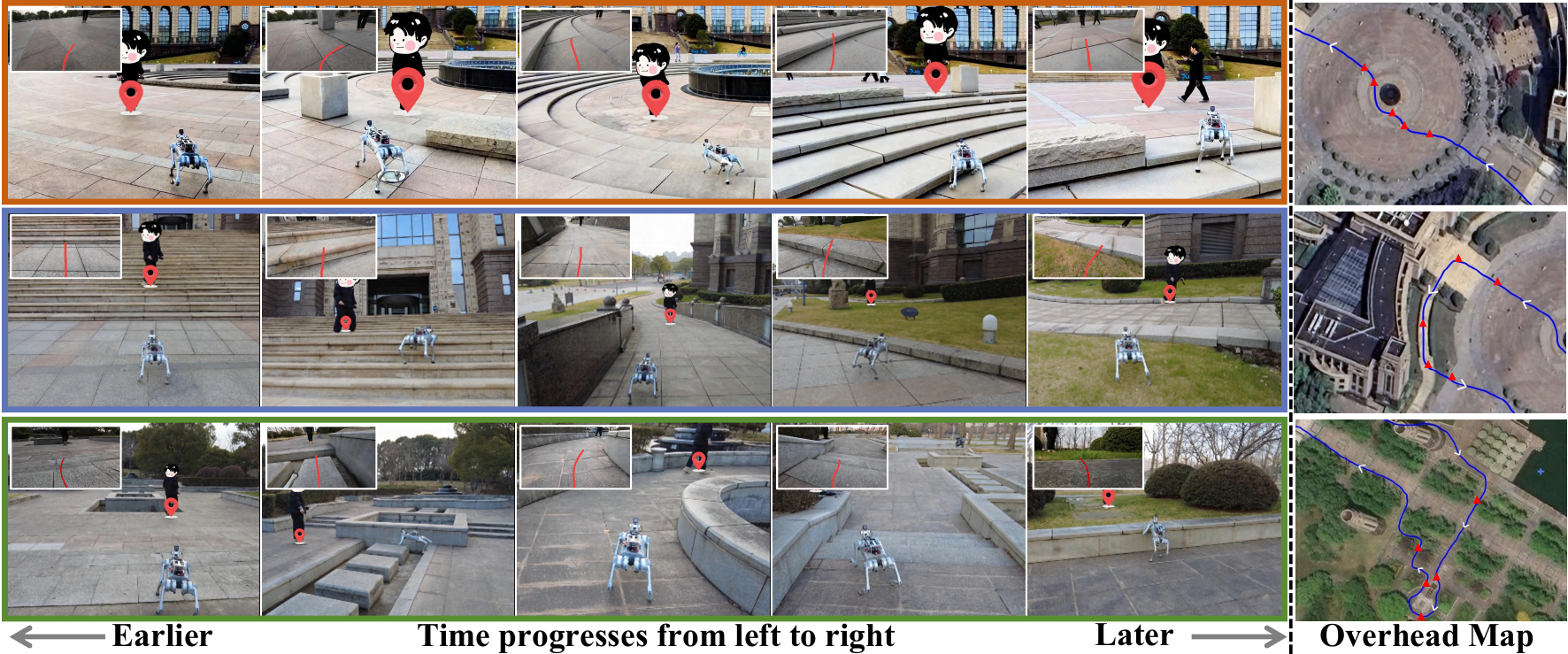}
    \caption{Long-distance outdoor dynamic target-following experiment. We conduct a 1-km outdoor experiment across diverse terrains. \textbf{\textcolor[rgb]{0.776,0.372,0.063}{Orange box}} indicates traversal of a circular plaza with flat ascending and descending stairs. \textbf{\textcolor[rgb]{0.368,0.459,0.729}{Blue box}} denotes climbing steep stairs followed by descent, grass traversal, and obstacle avoidance. \textbf{\textcolor[rgb]{0.345,0.557,0.192}{Green box}} represents gap crossing, traversal of a circular garden, and ending with a hurdle jump. The right panel shows the overhead navigation map.}
    \label{outdoor}
    \vspace{-0.5cm}
\end{figure*}

To evaluate the long-range fixed-goal navigation capability of MulDP, as shown in Fig.~\ref{sim}, we construct a 16 m × 16 m large-scale environment in Isaac Sim that is unseen during training. The fixed goal, marked by a red sphere, is located 13 m from the start. Under identical start–goal conditions, sampling from the solution space modeled by MulDP produces diverse and feasible navigation trajectories, reflecting the multimodal nature of the solution space. Unlike deterministic regression policies that converge to an averaged behavior, MulDP models a distribution over feasible motion plans via diffusion, enabling safe and efficient parkour navigation.

\begin{figure}[htbp]
\centerline{\includegraphics[width=\columnwidth]{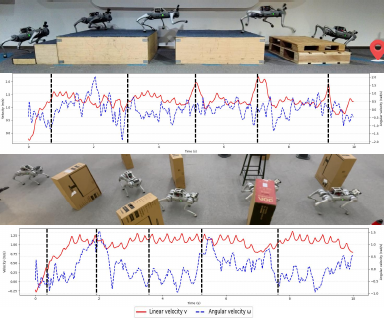}}
\caption{Velocity of the robot equipped with MulDP. \textbf{Black lines} indicate the instantaneous linear and angular velocities.}
\label{v}
\vspace{-0.8cm}
\end{figure}

Additionally, as shown in Fig.~\ref{headfig}, we evaluate the fixed-goal navigation capability of MulDP in the real world. Fig.~\ref{headfig} (left) illustrates the diversity of navigation plans generated by MulDP in the presence of static obstacles. Fig.~\ref{headfig} (bottom) demonstrates its safety and effectiveness in scenarios involving dynamic obstacles, demonstrating its ability to perform instantaneous perception, real-time motion planning, and diversified navigation.

As shown in Fig.~\ref{v}, the velocity profiles exhibit clear phase-dependent patterns aligned with terrain interaction events (dashed vertical lines). In the parkour scenario, the robot accelerates before reaching gap edges, with peak velocities approaching 2 m/s, and reduces speed only after landing. This behavior indicates that the policy anticipates upcoming terrain discontinuities and actively accumulates momentum for successful traversal, rather than reacting only upon contact. In the dense obstacle scenario, the robot maintains a relatively high and stable linear velocity (average around 1.1 m/s) while adjusting angular velocity to avoid obstacles. When encountering obstacles, the angular velocity reaches local peaks while the linear velocity remains smooth. This suggests that MulDP prefers heading adjustment over abrupt deceleration, favoring dynamically consistent avoidance instead of conservative stop-and-turn maneuvers. Such behavior preserves motion continuity and locomotion stability, and reduces unnecessary velocity fluctuations and kinetic energy dissipation.

\begin{figure}[t!]
\centerline{\includegraphics[width=\columnwidth]{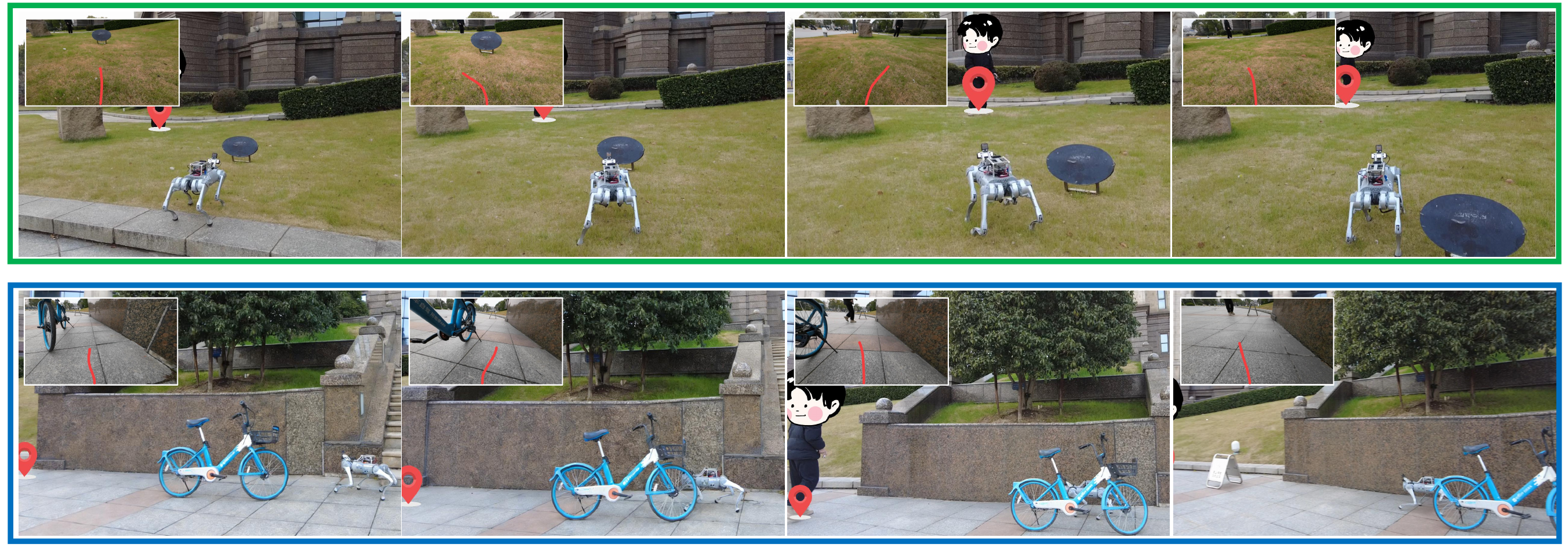}}
\caption{Atypical scenarios encountered in the dynamic target-following task, such as circular disc-shaped objects on grass (\textcolor[rgb]{0.345,0.557,0.192}{Green box}) and bicycles placed along the path (\textcolor[rgb]{0.368,0.459,0.729}{Blue box}).}
\label{outdoor_case}
\vspace{-0.6cm}
\end{figure}

\subsection{Dynamic Target-Following Navigation Task}


We conduct dynamic goal-tracking experiments in outdoor environments. In real-world settings, unavoidable disturbances such as illumination variations and motion blur introduce significant distribution shifts between the training data and onboard observations. Despite these challenges, MulDP preserves long-horizon decision consistency and dynamically adapts to evolving environments, demonstrating robust generalization. As shown in Fig.~\ref{outdoor}, it completes over 1 km of fully autonomous parkour navigation without human intervention, continuously tracking a dynamic target while executing diverse traversal behaviors, including stair ascent and descent, grass traversal, gap crossing, obstacle avoidance, and hurdle jumping.

Furthermore, as illustrated in Fig.~\ref{outdoor_case}, we evaluate the generalization capability of MulDP in outdoor scenarios involving uncommon obstacle geometries, such as circular disc-shaped objects on grass and bicycles placed along the path. These obstacle configurations are not explicitly encountered during training. Despite the appearance shift and irregular shapes, the robot accurately recognizes them as non-traversable structures and adapts its trajectory accordingly. Rather than attempting aggressive traversal behaviors, MulDP consistently opts for safe circumvention, demonstrating its ability to generalize obstacles beyond the training distribution. More qualitative results can be found in the accompanying supplementary video.

\section{CONCLUSIONS}

We propose a novel \textbf{Multimodal Diffusion Policy (MulDP)} that generates temporally coherent velocity-command horizons for autonomous quadruped parkour navigation. We also introduce the \textbf{Quadruped Parkour Navigation Dataset (QPND)}, to our knowledge the first multimodal dataset specifically designed for extreme quadruped parkour navigation, covering diverse terrains and rich navigation behaviors. By coupling depth perception, proprioception, goal information, and a low-level locomotion policy, MulDP moves beyond obstacle avoidance and enables traversal behaviors such as gap crossing, stair traversal, obstacle circumvention, and hurdle jumping. Simulation and real-world experiments demonstrate robust fixed-goal and dynamic target-following navigation, including long-distance outdoor deployment without human intervention. Future work will explore semantic perception, higher-level planning, and lightweight safety evaluation modules.


\bibliographystyle{IEEEtran}
\bibliography{refs}

\end{document}